\documentclass[conference]{IEEEtran}
\IEEEoverridecommandlockouts
\PassOptionsToPackage{numbers, compress}{natbib}
\usepackage{cite}
\usepackage{amsmath,amssymb,amsfonts}
\usepackage{algorithmic}
\usepackage{graphicx}
\usepackage{textcomp}
\usepackage{xcolor}
\usepackage{times}
\usepackage{epsfig}
\usepackage{bbm}
\usepackage{url}            
\usepackage{booktabs}      
\usepackage{nicefrac}
\usepackage{microtype}

\DeclareMathOperator*{\argmin}{arg\,min}

\def\BibTeX{{\rm B\kern-.05em{\sc i\kern-.025em b}\kern-.08em
    T\kern-.1667em\lower.7ex\hbox{E}\kern-.125emX}}

\graphicspath{ {./Figures/} }
\begin{document}

\title{BlurNet: Defense by Filtering the Feature Maps}

\author{\IEEEauthorblockN{Ravi S. Raju}
\IEEEauthorblockA{\textit{Department of Electrical Engineering} \\
\textit{University of Wisconsin-Madison}\\
Madison, WI, USA \\
rraju2@wisc.edu}
\and
\IEEEauthorblockN{Mikko Lipasti}
\IEEEauthorblockA{\textit{Department of Electrical Engineering} \\
\textit{University of Wisconsin-Madison}\\
  Madison, WI USA \\
  mikko@engr.wisc.edu}
}

\maketitle

\begin{abstract}
Recently, the field of adversarial machine learning has been garnering attention by showing that state-of-the-art deep neural networks are vulnerable to adversarial examples, stemming from small perturbations being added to the input image. 
  Adversarial examples are generated by a malicious adversary by obtaining access to the model parameters, such as gradient information, to alter the input or by attacking a substitute model and transferring those malicious examples over to attack the victim model.
  Specifically, one of these attack algorithms, Robust Physical Perturbations ($RP_2$), generates adversarial images of stop signs with black and white stickers to achieve high targeted misclassification rates against standard-architecture traffic sign classifiers.
  In this paper, we propose BlurNet, a defense against the $RP_2$ attack.  
  First, we motivate the defense with a frequency analysis of the first layer feature maps of the network on the LISA dataset, which shows that high frequency noise is introduced into the input image by the $RP_2$ algorithm. 
  To remove the high frequency noise, we introduce a depthwise convolution layer of standard blur kernels after the first layer. 
  We perform a blackbox transfer attack to show that low-pass filtering the feature maps is more beneficial than filtering the input.
  We then present various regularization schemes to incorporate this low-pass filtering behavior into the training regime of the network and perform white-box attacks. 
  We conclude with an adaptive attack evaluation to show that the success rate of the attack drops from 90\% to 20\% with total variation regularization, one of the proposed defenses.

\end{abstract}


\section{Introduction}
Machine learning has been ubiquitous in various fields like computer vision and speech recognition. \cite{Krizhevsky:2012:ICD:2999134.2999257, DBLP:journals/corr/abs-1303-5778} 
However, despite these advancements, neural network classifiers have been found to be susceptible to so called adversarial images \cite{intProps}. 
These images are created by altering some pixels in the input space so that a human cannot distinguish it from a natural image but a deep neural network will misclassify the input \cite{Linear}. 
This obviously has severe implications considering the rise of self-driving cars and computer vision systems being installed in industrial applications. 


To this end, we are interested in exploring a defense to the attack proposed in Robust Physical-World Attacks\cite{DBLP:journals/corr/EvtimovEFKLPRS17}.
In that work, the authors designed a general attack algorithm, Robust Physical Perturbations ($RP_2$) to generate visual adversarial perturbations which are supposed to mimic real-world obstacles to object classification.
They sample physical stop signs from varying distances and angles and use a mask to project a computed perturbation onto these images. 
On a standard classifier for road signs, their attack is 100\% successful in misclassifying stop signs.

\begin{figure}[t]
\includegraphics[scale=0.6]{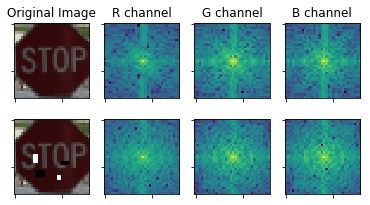}
\caption{The frequency spectrum of an unperturbed and perturbed stop sign with a sticker attack. The spectrum has been log-shifted and normalized. Frequencies close to the center correspond to lower frequencies and those that are near the edges correspond to higher ones. Observing the spectrum of the stop sign does not give a clear indication where the perturbations from the stickers lie. Yellow corresponds to regions with most information content.}
\label{input_blurring}
\end{figure}

Many defenses use spatial smoothing\cite{DBLP:journals/corr/LiL16e,xu2017feature,DBLP:journals/corr/LiangLSLSW17} as means to stamp out the perturbation caused by adversarial attacks. Unfortunately this approach is not always effective if the perturbation is in the form of a piece of tape on a stop sign.
To verify this, we plot the Fast Fourier Transform (FFT) spectrum of a vanilla and perturbed stop sign in Figure \ref{input_blurring}. 
Qualitatively, there does not seem to be any significant difference between the two spectra making filtering the input a questionable defense.

In this paper, we explore introducing low-pass filtering to the feature maps in the first layer of the network.
The main idea is to curb the spikes in the feature map caused by the perturbations by convolving them with a standard blur kernel or via regularization. 
This will squash the spikes at the expense of attenuating the signal at the output layer. 
We begin by giving an overview of the $RP_2$ algorithm and some background on machine learning security. 
In section 3, we discuss details of adding a filtering layer to dampen high frequency perturbations and perform a blackbox evaluation of $RP_2$ compared with input filtering.
In section 4, we propose various regularization schemes for the network to learn the optimal parameters to incorporate the low-pass filtering behavior: using the $L_{\infty}$ norm, total variation minimization \cite{Rudin:1992:NTV:142269.142312}, and Tikhonov regularization using smoothing regularization operators \cite{article}.
We then perform a white-box evaluation with $RP_2$ and find that the best performing algorithms are total variation regularization and Tikhonov regularization, which reduce the attack success rate from 90\% to 17.5\% and 10\% respectively compared to the baseline classifier.
In section 5, we test the upper bounds of each of the defense methods proposed with a specialized adaptive attack for each defense. 
We observed that the robustness for the Tikhonov regularization dropped by 30\% and that the total variation regularization is the truly robust solution with the attack success rate capped at 20\%.

\section{Background}
\subsection{Problem definition}
Consider a neural network to be a function $F(x) = y$, such that $F: \mathbb{R}^{m} \rightarrow \mathbb{R}^{n}$, where $x$ is the input image and $m = h * w * c$ such that $h, w, c$ are the image height, image width, and channel number and $n$ is a class probability vector of length of number of classes denoting the class of the image. The goal of an attack algorithm is to generate an image, $x_{adv}$, so the classifier output mislabels the input image in this manner, $F(x_{adv}) \neq y$. The attack success rate is defined as the number of predictions altered by an attack, that is, $\frac{1}{N}\sum_{n=1}^{N} \mathbbm{1}[F(x_n) \neq F(x_{nadv})]$. Another metric to characterize the attacker's success is the dissimilarity distance between natural and adversarial images,
$$
  \frac{1}{N}\sum^{N}_{n=1} \frac{||x - x_{adv}||_{p}}{||x||_{p}}.
$$
where \textit{p} can take different values depending on the norm chosen; we restrict ourselves to the $L_2$ case. An adversarial attack is considered strong if the attack success rate is high while having a low dissimilarity distance.

\subsection{Robust Physical Perturbations Attack}
We provide a description of the threat model that was considered when developing our defense. This algorithm is restricted to the domain of road sign classification, focused on finding effective physical perturbations that are agnostic to unconstrained environmental conditions such as distance and the viewing angle of the sign. 
This is called the Robust Physical Perturbation Attack, known as $RP_2$.
$RP_2$ is formulated as a single-image optimization problem which seeks to find the optimal  perturbation $\delta$ to add to an image $x$, such that the perturbed input 
$x' = x + \delta$ causes the target classifier, 
$f_\theta(\cdot)$ to incur a misclassification: 
$\min H(x + \delta, x), s.t. f_\theta(x + \delta) = y^{*}$
where $H$ is a chosen distance function and $y^{*}$ is the target class. 
\par
To solve the above constrained optimization problem, it must be reformulated in Lagrangian-relaxed form \cite{CWattack}. This threat model differs from all the others\cite{Linear,CWattack,DBLP:journals/corr/KurakinGB16} in that the noise introduced must be concentrated on a specific region of image. In the context of road sign classification, an attacker cannot alter the background of the scene so is therefore constrained to the sign itself. To mimic this effect, a binary mask, $M_x$, is multiplied with the perturbation, $\delta$, to concentrate the perturbation onto the sign. Since the perturbation is printed in the real world, we need to account for fabrication error, which ensures that the perturbation produced is a valid color in the physical world. This is quantified by the non-printability score(NPS), defined by Sharif \textit{et al.} \cite{sharif2016accessorize} given by: 
$$ NPS = \sum_{\hat{p} \in R(\delta)}  \prod_{p\sp{\prime} \in P} | \hat{p} - p\sp{\prime} |,$$ where $P$ is a set of printable colors and $R(\delta)$ is the set of RGB triples used in the perturbation.
\par
The final formulation of the optimization of the perturbation is presented as follows: 
\begin{align}
\argmin_{\delta} \lambda\lVert M_x \cdot \delta \rVert_p + NPS + \nonumber \\ J(f_\theta(x_i + T_i(M_x \cdot \delta)), y^*).
\end{align}
Here, $T_i$ is an alignment function for the masked perturbation that is used to account for if the image, $x_i$, was transformed so it can be placed optimally; $J$ is the cross-entropy loss of the classifier. For the distance metric in $||\cdot||$, both the $L_1$ and $L_2$ norms can be considered. More details on the algorithm can be obtain from \cite{DBLP:journals/corr/EvtimovEFKLPRS17}.

\subsection{Transferability and Adaptive Attacks }
Another aspect of these adversarial attack algorithms is the \textit{transferability property}.
The idea behind transferability is that adversarial examples that are generated from a model where all the parameters are known, can be transferred over to another model that is not known to the attacker. 
It has been shown these transferability attacks can be performed between different classes of classifiers such as deep neural networks(DNNs), SVMs, nearest neighbors, etc. \cite{DBLP:journals/corr/PapernotMG16}
The motivation for black box attack models arises from this property wherein the adversary is aware of the defense being deployed but does not have access to any of the network parameters or the exact training data \cite{DBLP:journals/corr/CarliniW17}.  
This is the most difficult threat setting for the adversary to operate under as opposed to a white-box setting, in which all the information about the model parameters are known.
According to Athalye \textit{et al.} \cite{DBLP:journals/corr/abs-1802-00420,DBLP:journals/corr/abs-1902-06705}, in order to evaluate any defense that lacks provable guarantees, it is necessary to modify the attack algorithm so that the defense's effectiveness is tested against new attacks under the specified threat model. 
This can be done by modifying the attacker's loss function.
In this paper, for every new defense method we explore, we attempt an adaptive attack to capture the true robustness of the defense.

\subsection{Dataset and Model}
We adopt the setup from \cite{DBLP:journals/corr/EvtimovEFKLPRS17} by examining the LISA dataset \cite{LISA} and a standard 4 layer DNN classifier in the Cleverhans framework \cite{cleverhans}. LISA is a standard U.S traffic sign dataset containing 47 various signs, but since there exists a large class imbalance, we only consider the top 18 classes, just as \cite{DBLP:journals/corr/EvtimovEFKLPRS17}. The network architecture is comprised of 3 convolution layers and a fully-connected layer. We train all the classifiers for 2000 epochs with the ADAM optimizer with $\beta_1 = 0.9$, $\beta_2 = 0.999$, and $\epsilon = 10^{-8}$. We evaluate our defense based on a sample set of 40 stop sign images provided by \cite{DBLP:journals/corr/EvtimovEFKLPRS17} in their github repo. 

\section{Motivation}
We begin by analyzing the effects of adding a sticker via the $RP_2$ algorithm via observations of the feature maps of the classifier. 
Understanding differences in activations in both natural and adversarial examples can inform the design of an appropriate defense strategy.
When we visualize the feature maps, we can observe an unwanted spike from the activation maps from the first layer in the spatial location where the mask is inserted over the sign. 
These spikes are large enough that as the activations propagate through the network they cause the classifier to misclassify the input \cite{DBLP:journals/corr/ZantedeschiNR17,Linear}.
\begin{figure}[t]
\centering
\includegraphics[scale=0.5]{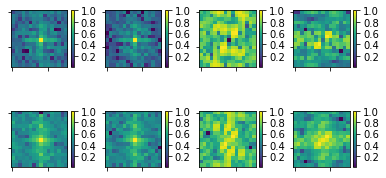}
\caption{The FFT Spectrum of a subsampling of feature maps from the first layer of the network. Each row corresponds to one unique feature map. The first column is the spectrum of feature maps of an unperturbed stop sign. The second column corresponds to the spectrum of feature maps of a stop sign with the sticker attached. The third column is a difference between the unperturbed and perturbed spectrum. Finally, the fourth column is a blurred version of the difference spectrum. Values were normalized.}
\label{fft_feat_maps}
\end{figure}
Based on the assumptions of the threat model, the perturbation is constrained to be on the stop sign, which suggests that the neighboring values around the region of the perturbation are dissimilar in the activation map. 
In general, we would normally expect smooth transitions in the activation map of images; that is, neighbor activations within some spatial window should have approximately similar values.
As a motivating example, we applied a standard 5x5 low-pass blur kernel over each of the feature maps. 
As a result of applying the filter the impact of the spike was substantially smaller.

This initial analysis motivates us to propose a defense that applies a set of low-pass filters to the output of first layer of the network. 
For isolated spikes that are caused by adversarial perturbations, low-pass filters are a natural fit for smoothing out unexpected spatial transitions in the activation maps. 
We focus exclusively on the feature maps after the first layer since spatial locality of the perturbation is still preserved. 
We insert these filters by performing a depthwise convolution on the feature maps to ensure that the filters are applied independently to each channel \cite{DBLP:journals/corr/Chollet16a}.

To evaluate the efficacy of inserting the depthwise convolution layer, we transform the feature maps into the frequency domain by computing the Fast Fourier Transform (FFT) of the natural image, adversarial image, and their respective blurred images, as shown in Figure \ref{fft_feat_maps}. 
The spectrum is on a log-scale and shifted so points close to the center correspond to lower frequencies and points near the edges to higher frequencies. 
Based on Figure \ref{fft_feat_maps}, most of the high frequency artifacts introduced from the perturbations were removed.
We do observe some low-frequency components that were induced by the attack, but the influence from these, compared to the high frequency spikes, is much lower.

\begin{table}[h!]
  \begin{center}
    \caption{Results from black box evaluation}
    \label{tab:transfer}
    \begin{tabular}{c|c|c} 
      & \textbf{Accuracy} & \textbf{Attack Success Rate}\\
      \hline
      Baseline & 100\% & 90\%\\
      Input filter 3x3 & 100\% & 87.5\%\\
      Input filter 5x5 & 100\% & 67.5\%\\
      3x3 filter on L1 maps & 100\% & 65\%\\
      5x5 filter on L1 maps & 87.5\% & 17.5\%\\
    \end{tabular}
  \end{center}
\end{table}

\subsection{Filtering the Input vs. Filtering the feature map}
A natural question to ask is what is the motivation for filtering the feature maps over applying the blur kernel over the input image. 
Our hypothesis is that filtering at the input layer does not remove the perturbation at similar window sizes as filtering the feature maps.
To test this hypothesis, we perform a black box transferability attack by generating adversarial examples on the vanilla network and transferring them over to the same model with a blurring filter at the input and one on the feature maps.
For our transfer attack, we evaluate the accuracy on a subset of the unperturbed natural stop signs and then measure the attack success rate on the perturbed stop signs with $\lambda=0.002$ and ran for 300 epochs. 
Our results are shown in Table \ref{tab:transfer}. 
As seen in Table \ref{tab:transfer}, for lower kernel sizes, while blurring the input does not have much of an impact on the accuracy, it is not effective in alleviating the noise introduced by $RP_2$. 
Compared to blurring the input, adding the blur kernels to the features generated by the first layer seems to effectively reduce the attack success rate at some cost in accuracy.
This result motivates an attempt to alter the training regime so it learns the gain parameters in the filter implicitly, rather than setting them to predefined known values, so that robustness can be maintained at a minimal accuracy loss.

\subsection{Evaluating under a different threat model}
Before proceeding onto the white box evaluation that we present in Section IV, we are interested in seeing the performance of the defenses under the standard $\epsilon$-bound pixel-based adversaries.
We choose the traditional PGD attack\cite{PGD} as our adversary, with $\epsilon=8.0/255$, step size, $\alpha=0.01$ and 10 steps. The results are reported in Table \ref{tab:pgd_results} in the Supplementary Material. 
Unsuprisingly, all the defenses are broken under the assumption that the adversary can manipulate any arbitrary pixel since our defenses rely on the perturbation being localized on the subject and that neighboring features should be similar to each other. 
This finding suggests that there does not exist one broad defense to all attacks but rather suggests that defenses should be tailored to defend a restricted class of attacks instead.

\section{Learning the Filter Parameters}
From the previous section, we see that filtering the feature maps is an effective scheme at discarding the perturbations introduced from $RP_2$. However, the side effect of naively inserting a layer of low-pass filters is that the confidence of the prediction is reduced. In certain application domains such as autonomous vehicles, low confidence predictions from the classifier may not be acceptable. 
For correcting the reduction in the confidence, we seek to incorporate an additional loss term into the training of the classifier. 
We explore several different options for loss terms: $L_\infty$ norm on the weights of the depthwise layer (added filter layer), total variance(TV) minimization applied to the feature map (no added layer) \cite{Rudin:1992:NTV:142269.142312}, and Tikhonov regularization \cite{article}.
\subsection{Depthwise Convolution Layer}
To emulate the effect of adding a layer of low pass filters, the $L_\infty$ norm is an apt choice for the depthwise weights. This will ensure that the weights in the kernel take similar values to act much like a low pass filter. The resulting loss that is minimized, where K is the number of channels in the input, is: 
\begin{equation}
\begin{aligned}
\min \: \alpha\sum_{j = 1}^{K}\lVert W_{depthwise}[:,:,j] \rVert_{\infty} + J(f_\theta(x, y)).
\end{aligned}
\end{equation}
In our white-box evaluation, we explore the effect of filter width on the effectiveness of the attack algorithm.
\begin{table*}[t!]
  \begin{center}
    \caption{Results from white box evaluation}
    \label{tab:table2}
    \begin{tabular}{c|c|c|c|c|c} 
      & \textbf{$\alpha$} & \textbf{Legitimate Acc.} & \textbf{Average Success Rate} & \textbf{Worst Success Rate} & \textbf{$L_2$ Dissimilarity}\\
      \hline
      Baseline & 0 & 91\% & 49.18\% & 90\% & 0.207\\
      Gaussian aug ($\sigma$ = 0.1) & -& 84.3\%& 19.44\%& 62.5\%& 0.238\\
      Gaussian aug ($\sigma$ = 0.2) & -& 84.4\%& 55.97\%& 80\%& 0.196\\
      Gaussian aug ($\sigma$ = 0.3) & -& 85.6\%& 21.39\%& 25\%& 0.198\\
      Rand. sm ($\sigma$ = 0.1) & -& 84.3\%& 19.3\%& 67.5\%& 0.236\\
      Rand. sm ($\sigma$ = 0.2) & -& 84.4\%& 55\%& 70\%& 0.189\\
      Rand. sm ($\sigma$ = 0.3) & -& 85.6\%& 22.5\%& 22.5\%& 0.198 \\
      Adv-train & - & 77.9\% & 11.94\% & 20\% & 0.244 \\
      3x3 conv & $10^{-5}$ & 86.3\% & 30\% & 55\% & 0.201\\
      5x5 conv & 0.1 & 86.3\% & 24.11\% & 47.5\% & 0.189\\
      7x7 conv & 0.1 & 87\% & 11.61\% & 30\% & 0.203\\
      TV & $10^{-4}$ & 85.6\% & 7.92\% & 17.5\% & 0.224\\
      TV & $10^{-5}$ & 82.3\% & 8.47\% & 30\% & 0.199\\
      Tik$_{hf}$ & $10^{-4}$ & 84.5\% & 5.416\% & 10\% & 0.214\\
      Tik$_{pseudo}$ & $10^{-6}$ & 83.6\% & 13.9\% & 35\% & 0.222
    \end{tabular}
  \end{center}
\end{table*}

\subsection{Total Variation Loss Term}
We introduce the TV loss term into the optimization algorithm for the classifier, without adding an additional depthwise convolution to the network. Total variation of the image measures the pixel-level deviations for the nearest neighbor and minimizes the absolute difference between those neighbors. For a given image, the TV of an input image $x$ is given as: 
\begin{equation}
\begin{aligned}
TV(x) = \sum_{i,j} |x_{i + 1, j} - x_{i,j}| + |x_{i, j + 1} - x_{i,j}|.
\end{aligned}
\end{equation}
We omit the depthwise convolution layer from the network and instead let the first layer of the network learn to filter out the high-frequency spikes in its feature map. 
Defining $\mathcal{F}$ as the set of feature maps after the first layer, the final optimization objective is given as:
\begin{equation}
\begin{aligned}
\min \: \alpha_{TV} \frac{1}{N \cdot K}\sum_{i = 1}^{N}\sum_{i = 1}^{K} TV(\mathcal{F}[i,:,:,k]) + J(f_\theta(x, y)),
\end{aligned}
\end{equation}
where $N$ and $K$ are the batch size and the number of output channels, respectively.
Intuitively, TV removes effects of details that have high gradients in the image, effectively targeting the perturbations introduced by $RP_2$ for denoising. 
TV encourages the neighboring values in the feature maps to be similar so the high spike introduced by $RP_2$ would be diminished.

\subsection{Tikhonov Regularization}
In our last method, we propose two variants based off of Generalized Tikhonov Regularization. 
The general form for Tikhonov is given usually given as
\begin{equation}
\begin{aligned}
\min \: \mu ||L\cdot w_0||_2 + J(f_\theta(x, y, w_0)),
\end{aligned}
\end{equation}
where $w_0$ is some parameter of the model, $\mu$ is the regularization parameter, and $L$ is the regularization operator \cite{article} and is the identity matrix in the $L_2$ regularization case.
Similar to the TV loss terms, we attempt to induce low-pass filtering by applying regularization to the first layer feature maps. 
We do this by selecting a regularization operator $L$ which penalizes high frequency components in the feature maps. 
The two operators are as follows:
\begin{itemize}
  \item The operator $L_{hf} = (I - L_{avg})$, where $L_{avg}$ is a matrix which transforms the input into its moving average and consequently, $L_{hf}$ extracts the high frequency components and gets minimized. 
  \item The operator $L_{diff}^{+}$, where $L_{diff}$ is a difference matrix which approximates a derivative operation. When the difference matrix approximates a derivative operation, its pseudoinverse approximates an integral operation and is thus a low pass filter \cite{article}. These are known as smoothing operators. We compute this pseudoinverse and perform an elementwise multiplication to apply it to the feature maps.
\end{itemize}

Borrowing notation from above, the actual objectives being minimized are given as:

\begin{align}
\min \: & \alpha_{hf} \frac{1}{N \cdot K}\sum_{i = 1}^{N}\sum_{i = 1}^{K} ||L_{hf}\cdot\mathcal{F}[i,:,:,k]||_2 \nonumber \\
& + J(f_\theta(x, y))
\end{align}

\begin{align}
\min \: & \alpha_{pseudo} \frac{1}{N \cdot K}\sum_{i = 1}^{N}\sum_{i = 1}^{K} ||L_{diff}^{+}\cdot\mathcal{F}[i,:,:,k]||_2 \nonumber \\ & + J(f_\theta(x, y))
\end{align}
Both of these will be referred to as Tik$_{hf}$ and Tik$_{pseudo}$ respectively. Similar to the other defenses, this regularization scheme should enforce the property that lower frequency representations should be more emphasized in the feature maps. To reiterate, there are no changes to the model architecture; the defense is coming from the first layer weights.
\subsection{White-Box Evaluation}

We perform a white-box evaluation and sweep the hyperparameters in the attack algorithm, $\lambda$ and the attack target, $y^{*}$. Our results are reported in Table \ref{tab:table2}. 
In the white-box setting, the attacker has access to all the model parameters as well as the classification output. 
The main goal of the evaluation is to detect if the attack algorithm is able to introduce perturbations with the knowledge of the model parameters.
The legitimate accuracy corresponds to the accuracy on the test set. 
We ran the attack algorithm for 300 epochs.
When we sweep the parameters, we find that the attack target is the parameter most sensitive to increasing the success rate of the attack and is relatively invariant to $\lambda$.
Certain attack targets are more amenable to attacks because there may be steeper gradients of the loss function with respect to those target labels. 
The legitimate accuracy refers to the accuracy on the test set and the average success rate is the attack success rate averaged across all other 17 classes (excluding the true label).
We also report the best case scenario for the attacker and the $L_2$ dissimilarity distance. 

We use Randomized Smoothing\cite{DBLP:journals/corr/abs-1709-05583} as a baseline to compare our low-pass filtering methods. 
In their paper, Cohen et. al \cite{DBLP:journals/corr/abs-1902-02918} leverage the classifier trained with Gaussian-augmented noise to obtain a new smoothened classifier which gives provable guarantees on the accuracy of the classifier. 
Intuitively, it can be interpreted that this augmentation is acting like a smoothing operation by drowning the adversarial perturbations, making it an appropriate method for baseline comparison. We took 100 MC samples when evaluating the forward prediction on the augmented images.
We also use standard PGD-Adversarial training as a baseline comparison with an $L_{\infty}$ adversary with $\epsilon=8/255$ and step size, $\alpha=0.1$ with 7 steps of gradient descent\cite{PGD}. For each epoch, we train on 50\% on clean examples and the other half on Adversarial examples. One potential explanation to why these traditional pixel-based defenses do not perform as well is that they were developed under a different threat model and were not designed to defend against this type of attack, which further supports the point no universal defense exists against all attacks.

We find that the Tik$_{hf}$ and TV minimization loss term has superior performance compared to all the other methods, bringing the attack success rate down to 10\% and 17.5\% respectively.
Tik$_{hf}$ is able to directly minimize high frequency components which is why the success rate is lower than the success rate for total variation regularization. 
TV is effectively encouraging the first convolutional weight to not only act as a feature extractor but also to stifle high variations coming from the input.
For the depthwise convolution layer with the $L_{\infty}$ norm regularizer, as the width of the filter increases, the  network is able to attenuate the attack success rate because a large window from the surrounding neighbors will be able to smooth out the perturbation.
However, the TV loss is better than applying the $L_{\infty}$ as it is directly able to influence the weights to behave like a low-pass filter rather than indirectly through the $L_{\infty}$ norm.
The Tik$_{pseudo}$ model has comparable performance to the 7x7 depthwise convolution. A plausible reason why Tik$_{pseudo}$ performs slightly worse is because it is an approximation of a low-pass filter, rather than the other methods which take a more direct approach against the high frequency components in the feature maps. 

\section{Adaptive Attacks}
After presenting the whitebox evaluation, we perform adaptive attacks for each of the defenses proposed to obtain an upper bound on the attack success rate. We do this by changing the loss function of the attacker to reflect what training or architectural changes we made to make the model more robust. We make sure to follow guidelines outlined in \cite{tramer2020adaptive} to show that we properly evaluate our defenses.

\begin{table*}[t!]
  \begin{center}
    \caption{Results from adaptive attack evaluation}
    \label{tab:adpt}
    \begin{tabular}{c|c|c|c|c|c}
      & \textbf{Average Success Rate} & \textbf{Worst Success Rate} & \textbf{$L_2$ Dissimilarity}\\
      \hline
      3x3 conv & 22.91\% & 52.5\% & 0.546\\
      5x5 conv & 46.25\% & 75\% & 0.539\\
      7x7 conv & 10.416\% & 20\% & 0.539\\
      TV ($10^{-4}$) & 8.33\% & 20\% & 0.044\\
      TV ($10^{-5}$) & 6.11\% & 25\% & 0.046\\
      Tik$_{hf}$ & 23.6\% & 47.5\% & 0.147\\
      Tik$_{pseudo}$ & 17.5\% & 45\% & 0.141\\
    \end{tabular}
  \end{center}
\end{table*}

\subsection{Low Frequency Attack on Depthwise Convolution}
We begin by trying to introduce low-frequency perturbations by restricting the $\delta$ parameter to a low-frequency region with a mask, $M_{dim}$. 
We follow Yash et al. \cite{DBLP:journals/corr/abs-1903-00073} by applying the Direct Cosine Transform (DCT) to the masked perturbation, 
\begin{align}
\argmin_{\delta} \: \lambda\lVert M_x \cdot \delta \rVert_p + J(f_\theta(x_i + \nonumber \\ T_i(IDCT(M_{dim}\cdot DCT(M_x \cdot \delta)))), y^*).
\end{align}
where the default dimension selected was 16. 
The results from the low frequency attack is report in Table \ref{tab:adpt}.
We observe that the adaptive attack on the 3x3 kernel did not significantly change, although the robustness of the 3x3 kernel was not as effective. The 5x5 kernel's performance was greatly diminished such that the worst case attack success rate jumped from 47.5\% to 75\%.  \par
\begin{figure}[b]
\centering
\includegraphics[scale=0.4]{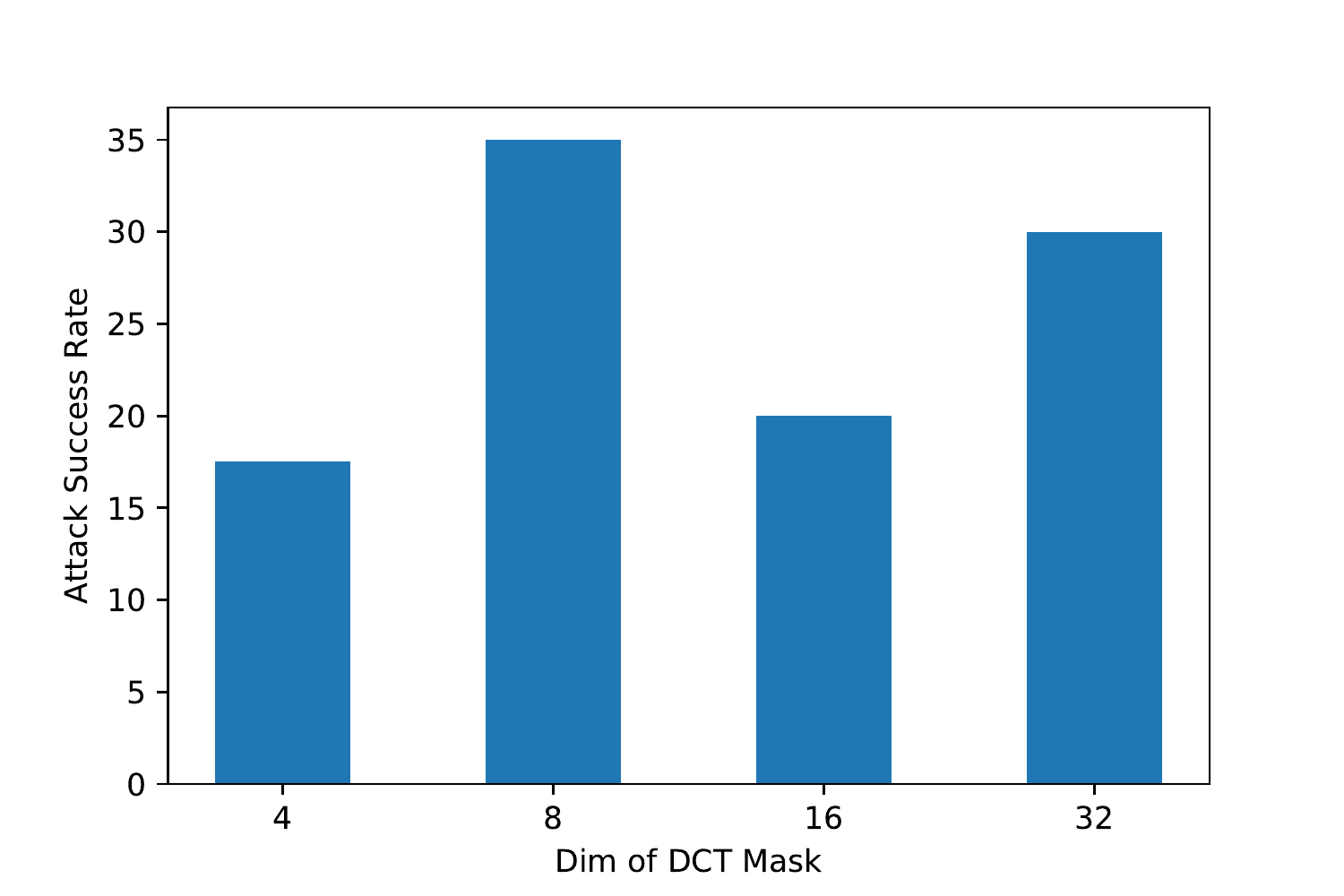}
\caption{The impact of the DCT mask dimension on the adaptive attack success rate for 7x7 depthwise convolution.}
\label{fft_dim_vs_ASR}
\end{figure}

In the 7x7 case, the attack did not produce an upper bound on the attack success rate. 
This could be due to the fact that the mask used was not restrictive enough for the 7x7 adaptive to succeed. 
In Figure \ref{fft_dim_vs_ASR}, we vary the dimension of $M_{dim}$ to try to increase the effectiveness of the attack.
We observe that when $M_{dim}$ is size 8x8, the attack success rate increases to 35\% showing that the adaptive attack is effective but not by a large margin showing that using $L_{\inf}$ regularized depthwise convolution is effective even against dynamic adversaries under the $RP_2$ threat model. When we observe the difference in the images between the white-box and adaptive attacks, we see that in the adaptive Adversarial examples the pixels in the sticker region tend to resemble values closer to their neighboring values. 

\subsection{Attack on Total Variation, Tikhonov Regularized Models}
For our adaptive attacks on our regularized models, we adapt the loss function of the attack to include details on how the model was trained. Specifically, we add the regularizer that the model was trained with to the attacker's loss function. We execute this attack on TV-regularized models, Tik$_{hf}$ model and the Tik$_{pseudo}$ model which are shown below. We abbreviate the previous loss terms in the attack objective as $l_{adv}$ for brevity.

\begin{align}
\argmin_{\delta} \: l_{adv} + \frac{1}{N \cdot K}\sum_{i = 1}^{N}\sum_{i = 1}^{K} TV(\mathcal{F}[i,:,:,k]),
\end{align}

\begin{align}
\argmin_{\delta} \: l_{adv} + \frac{1}{N \cdot K}\sum_{i = 1}^{N}\sum_{i = 1}^{K} ||L_{hf}\cdot\mathcal{F}[i,:,:,k]||_2,
\end{align}

\begin{align}
\argmin_{\delta} \: l_{adv} + \frac{1}{N \cdot K}\sum_{i = 1}^{N}\sum_{i = 1}^{K} ||L_{diff}^{+}\cdot\mathcal{F}[i,:,:,k]||_2.
\end{align}

We note the results of our adaptive attack in Table \ref{tab:adpt}.
We experimented with adding a regularization parameter to the additional term in the attack objective, but we found that this reduced the performance of the attack so we omitted these results.
From the white-box evaluation, it seemed that the most robust model was Tik$_{hf}$; however, upon conducting the adaptive attack, we can see the robustness for Tik$_{hf}$ drop by nearly 30\% whereas TV ($10^{-4}$) suffered only a 2.5\% performance degradation. Tik$_{pseudo}$'s worst case success rate was 10\% over the white-box evaluation. Clearly, the truly robust model is TV, maintaining low attack success rate even against a dynamic adversary. Under the $RP_2$ threat model, this result shows that performing filtering operations on the feature maps gains significant robustness with minimal accuracy loss. 

One of the possible reasons why Tikhonov regularization did not produce robust models could be due to the choice of the regularization operators. 
In the case of Tik$_{hf}$, increasing the effective averaging window of the operator $L_{avg}$ would be more aggressive in filtering, potentially at the cost of accuracy. For Tik$_{pseudo}$, \cite{article} discusses using a weighted pseudoinverse with the input (feature maps) may be a better approximation of a low pass filter. Intuitively, a superior approximation of the filter would be calculated with knowledge of the input rather than only a fixed regularization operator.

\section{Conclusion}
We performed spectral analysis of feature maps and saw that attacks introduce high-frequency components, which are amenable to low-pass filtering.
Our proposal introduced a simple solution of adding low-pass filters after the first layer of the DNN. 
We compare with this with blurring the input image and show that blurring at the feature level can confer some robustness benefit at the cost of some accuracy by performing a black-box transferability attack with $RP_2$.
To compensate for the loss in accuracy, we explore various regularization schemes: adding a depthwise convolution, total variation minimization, and Tikhonov regularization and show that we can recover the loss in accuracy while retaining significant robustness benefit under a white-box evaluation. Against the $RP_2$ attack, we showed that our defense can perform better than more traditional defenses against norm-bound adversaries such as \cite{PGD,DBLP:journals/corr/abs-1902-02918}.
We empirically show the upper bounds of these methods by performing an adaptive attack on the defense.
In the future, we hope to explore if BlurNet would still be robust against different kinds of physically realizable attacks.

\vspace{12pt}
{\small
\bibliographystyle{ieee_fullname}
\bibliography{reference}
}
\section{Supplementary Material}

\subsection{Inserting filters in higher layers}
\begin{figure}[h!]
\centering
\includegraphics[scale=0.55]{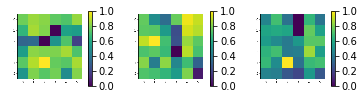}
\caption{The FFT Spectrum of a subsampling of feature maps from the second layer of the network. These feature maps were obtained from a normal stop sign. The high values indicated at the edges of the spectrum suggests that higher frequency information is relevant to maintain decent classification.}
\label{higher_feat_maps}
\end{figure}

We choose only to look at the feature maps after the output of the first layer.
We explored adding filters into higher layers of the neural network and we find that these reduce classification accuracy.
We hypothesize the reason for this accuracy loss is that the higher layers in the network naturally contain high frequency information. 
We verify this hypothesis by computing the Fast Fourier transform of the activations maps of the higher level convolutional layers given in Figure \ref{higher_feat_maps}.
From Figure \ref{higher_feat_maps}, we can see that the magnitude spectrum shows that the difference between higher and lower frequencies is not pronounced.
If a low pass filter is introduced at this level in the network, too much information is lost for the DNN to make a meaningful prediction. 
In order to maintain classification accuracy, high frequencies in the feature maps should not be squashed.
Adding a set of filters to the higher levels of the network is also difficult to justify from a semantic perspective, since the spatial locality of the features is not preserved, as the the receptive field of the neurons in upper layers is wider and even discontinuous due to non-unit convolution strides and/or max-pooling layers.
\subsection{Plots of the ASR vs. $L_2$ dissimilarity over all targets}
\begin{figure}[t]
\centering
\includegraphics[scale=0.45]{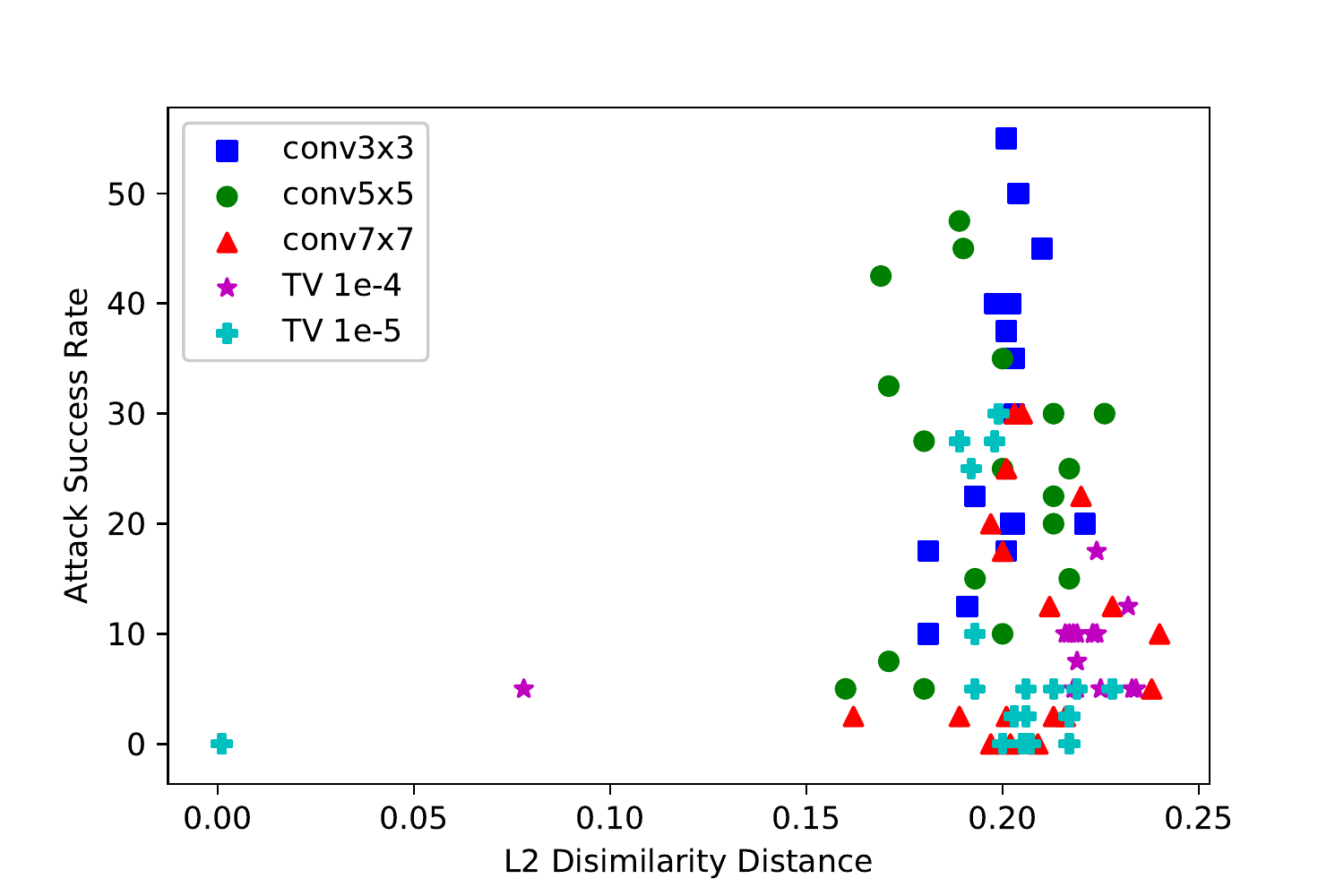}
\caption{Plot of the L2 Dissimilarity Distance against the Attack Success Rate for varying convolution widths and TV regularization. Lower and to the right is better.}
\label{L2_vs_atkplot}
\end{figure}

\begin{figure}[t]
\centering
\includegraphics[scale=0.45]{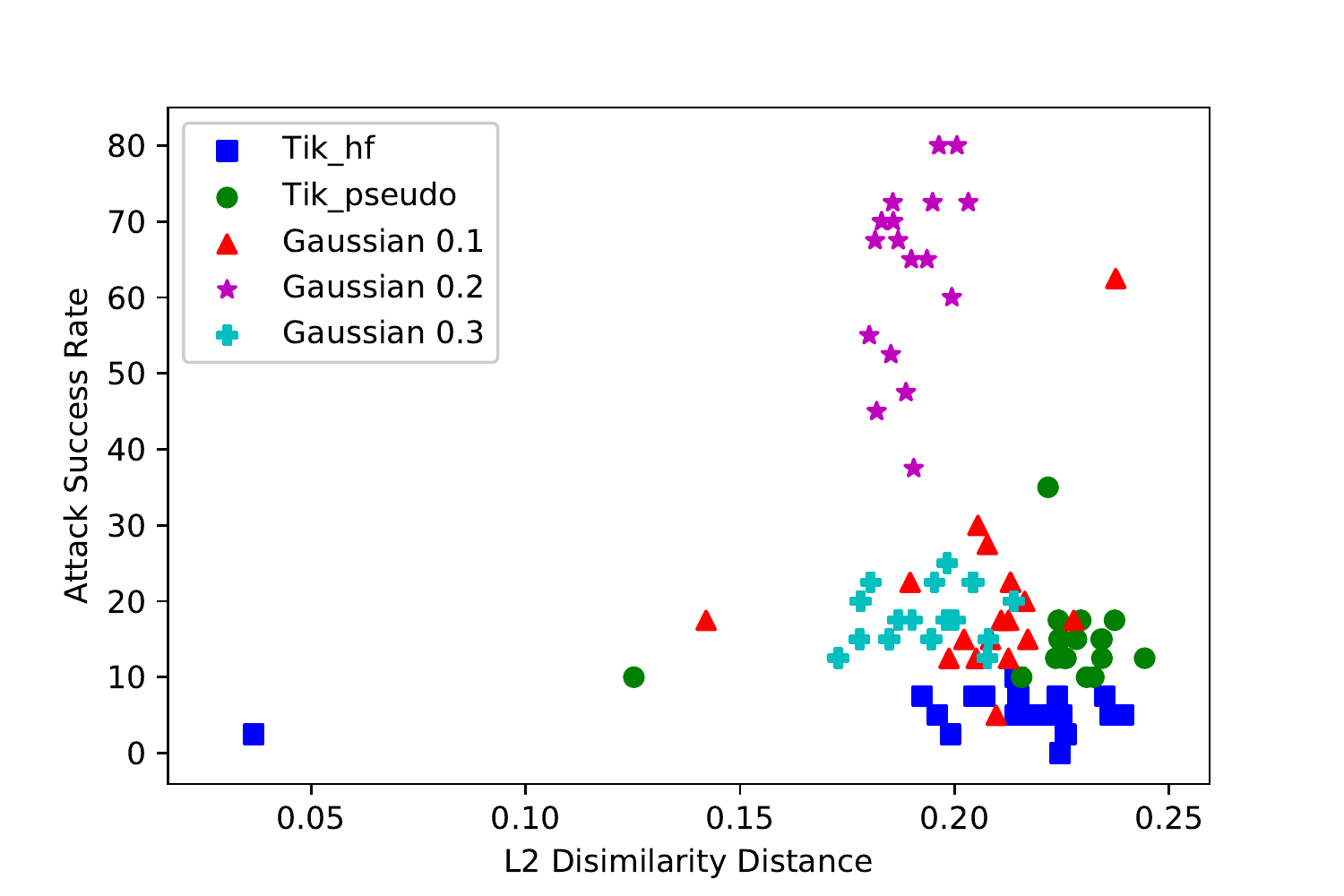}
\caption{Plot of the L2 Dissimilarity Distance against the Attack Success Rate for Tikhonov-regularized models and Gaussian augmentation. Lower and to the right is better.}
\label{L2_vs_atkplot_p2}
\end{figure}

In Figure \ref{L2_vs_atkplot} and \ref{L2_vs_atkplot_p2}, we plot the $L_2$ dissimilarity distance against the attack success rate to show the variation of each defense methods across the target labels. We find that TV and Tikhonov loss terms have less variation than the other depthwise convolution layers. TV and Tik$_{hf}$ proposed outperformed the Gaussian baseline both in terms of both average and worst case success rate and $L_2$ dissimilarity distance. 7x7 depthwise convolution and Tik$_{pseudo}$ were only outshown by 5\% by the best Gaussian augmented model ($\sigma = 0.3$) for the worse case attack success rate while having better average attack success rates and $L_2$ dissimilarity distance. Empirically, these results show performing smoothing and filtering operations on the feature maps rather than the input is more effective.

\subsection{Comparing Adversarial Training To Adaptive Attacks}
In Table \ref{tab:adveval}, we compare adversarial training against the TV adaptive, Tik$_{hf}$ and Tik$_{pseudo}$ attacks. Adversarial training is able to outperform all the defenses we proposed besides the TV-regularized defence. 
This evaluation shows that there is an advantage using total variation regularization in the case of the $RP_2$ attack over adversarial training, further reinforcing the point that no universal defense exists for all threat models.
We would also like to emphasize that performing adaptive attacks steps outside the RP$_{2}$ threat model in that manipulating input pixels falls outside the domain of stickers on a subject.

\subsection{Related Work}
Many kinds of defenses have been proposed in the machine learning security literature. 
There seems to have been two kinds of approaches to developing defenses: robust classification and detection. Robust classification refers to the classifier being able to correctly classify the input despite the perturbation whereas detection refers to a scheme of identifying if an example has been tampered with and rejecting it from the classifier. Recently, detection methods have seen much more popularity than robust classification (our method belongs to the latter class).
However, in certain domains such as autonomous vehicles, it is not always feasible to reject the input from classifier. 

\subsubsection{Adversarial training} \textit{Adversarial training} is the technique of injecting adversarial examples and the corresponding gold standard labels into the training set \cite{intProps,Linear,PGD}. The motivation of this methodology is that the network will learn the adversarial perturbations introduced by the attacker. The problem with adversarial training is that it doubles the training time of the classifier as new examples need to be generated. Moreover, as shown by Papernot \textit{et al.}, adversarial training needs all types of adversarial examples produced by all known attacks, as the training process is non-adaptive \cite{DBLP:journals/corr/PapernotMSW16}. Our method can be paired with any of these types of defenses. 
In their paper, Xie \textit{et al.} \cite{DBLP:journals/corr/abs-1812-03411} used feature denoising along with adversarial training to boost the performance of the network. While the spirit of the prior work is similar, some differences between this method and ours is that in our defense was designed under a different threat model in which we are focused on removing high frequency components in the first level feature maps.
We enforce this behavior explicitly by including it in the training process whereas in the prior work it was embedded into the model as an architectural element.
Since this method is paired with adversarial training, it is unclear on what  robustness of the denoising operation is providing by itself, making it difficult to compare to our method which does not involve any adversarial training.

\begin{table*}[h!]
  \begin{center}
    \caption{PGD results evaluation}
    \label{tab:pgd_results}
    \begin{tabular}{c|c|c}
      &\textbf{Attack Success Rate} & \textbf{$L_2$ Dissimilarity}\\
      \hline
      Baseline & 100\% & 0.53\\
      3x3 conv & 100\% & 0.512\\
      5x5 conv & 100\% & 0.502\\
      7x7 conv & 100\% & 0.511\\
      TV ($10^{-4}$) & 100\% & 0.455\\
      TV ($10^{-5}$) & 100\% & 0.437\\
      Tik$_{hf}$ & 100\% & 0.464\\
      Tik$_{pseudo}$ & 100\% & 0.443\\
    \end{tabular}
  \end{center}
\end{table*}

\begin{table*}
  \begin{center}
    \caption{Evaluating Adversarial Training Against Adaptive Adversaries}
    \label{tab:adveval}
    \begin{tabular}{c|c|c|c|c|c}
      & \textbf{Average Success Rate} & \textbf{Worst Success Rate} & \textbf{$L_2$ Dissimilarity}\\
      \hline
      TV adaptive attack & 5.85\% & 27.5\% & 0.046\\
      Tik$_{hf}$ attack & 17.6\% & 18\% & 0.148\\
      Tik$_{pseudo}$ attack & 15\% & 17.5\% & 0.150\\
    \end{tabular}
  \end{center}
\end{table*}

\subsubsection{Input transformations} Most previous work has applied some type of transform to the input image. 
In their paper, Guo \textit{et al.} use total variance minimization and image quilting to transform the input image. 
They use random pixel dropout and reconstruct the image with the removed perturbation \cite{DBLP:journals/corr/abs-1711-00117}.
Dziugaite \textit{et al.} examined the effects of JPEG compression on adversarial images generated from the Fast Gradient Sign Method (FGSM) \cite{JPEG_comp,Linear}.
They report that for perturbations of small magnitude JPEG compression is able to recover some of the loss in classification accuracy, but not always. 
Xu \textit{et al.} introduce feature squeezing, a detection method based on reducing the color bit of each pixel in the input and spatially smoothing the input with a median filter \cite{xu2017feature}. 
In their paper, Li \textit{et al.} propose detecting adversarial examples by examining statistics from the convolutional layers and building a cascade classifier. They discover that they are able to recover some of the rejected samples by applying an average filter \cite{DBLP:journals/corr/LiL16e}.
Liang \textit{et al.} looked at using image processing techniques such as scalar quantization and a smoothing spatial filter to dampen the perturbations introduced. The authors introduce a metric, which they define as image entropy, to use different types of filters to smooth the input \cite{DBLP:journals/corr/LiangLSLSW17}.
We stress that the key difference between these approaches and the proposed methods is that we focus on applying these smoothing techniques through model changes and regularization on the feature maps rather than the input.

\subsubsection{Other Defenses}
Gradient masking refers to the phenomenon of the gradients being hidden from the adversary by reducing model sensitivity to small changes applied to the input \cite{DBLP:journals/corr/PapernotMGJCS16}. 
These can be due to operations that are added to the network that are not differentiable so regular gradient based attacks are insufficient.
Another class of gradient masking includes introducing randomization into the network. Stochastic Activation Pruning essentially performs dropout at each layer where nodes are dropped according to some weighted distribution \cite{DBLP:journals/corr/abs-1803-01442}.
Xie \textit{et al.} propose a randomization in which the defense randomly rescales and randomly zero-pads the input to an appropriate shape to feed to the classifier \cite{DBLP:journals/corr/abs-1711-01991}.
However, as Athalye \textit{et al.} have shown in their paper, gradient masking is not an effective defense since the adversary can apply the Backward Pass Differential Approximation attack, in which the attacker approximates derivatives by computing the forward path and backward path with an approximation of the function. 
Even against randomization, the authors introduce another attack, Expectation over Transformation (EOT), where the optimization algorithm minimizes the expectation of the transformation applied to the input \cite{DBLP:journals/corr/abs-1802-00420}. 
\end{document}